\documentclass[10pt,twocolumn,letterpaper]{article}

\usepackage{iccv}
\usepackage{times}
\usepackage{epsfig}
\usepackage{amsmath}
\usepackage{amssymb}

\usepackage[utf8]{inputenc} 
\usepackage[T1]{fontenc}    
\usepackage{url}            
\usepackage{booktabs}       
\usepackage{amsfonts}       
\usepackage{nicefrac}       
\usepackage{microtype}      
\usepackage{xcolor}         

\usepackage{multirow}
\usepackage{booktabs}
\usepackage{verbatim}
\usepackage{subfig}
\usepackage{bm}
\usepackage{adjustbox}
\usepackage{footnote}
\makesavenoteenv{tabular}
\makesavenoteenv{table}

\usepackage[font=small,labelfont=bf]{caption}  
\usepackage{makecell}
\usepackage{tabulary}
\newlength\savewidth\newcommand\shline{\noalign{\global\savewidth\arrayrulewidth
  \global\arrayrulewidth 1pt}\hline\noalign{\global\arrayrulewidth\savewidth}}

\newcommand{\tablestyle}[2]{\setlength{\tabcolsep}{#1}\renewcommand{\arraystretch}{#2}\centering\footnotesize}

\newcolumntype{C}[1]{>{\centering\arraybackslash}p{#1}}
\newcolumntype{R}[1]{>{\raggedleft\arraybackslash}p{#1}}
\newcolumntype{L}[1]{>{\raggedright\arraybackslash}p{#1}}

\newcommand{\newcelll}[2][l]{%
  \begin{tabular}[#1]{@{}l@{}}#2\end{tabular}}
  \newcommand{\newcellr}[2][r]{%
  \begin{tabular}[#1]{@{}r@{}}#2\end{tabular}}

\makeatletter
\@namedef{ver@everyshi.sty}{}
\makeatother
\usepackage{tikz}
\newcommand{\Arrow}[1]{%
\parbox{#1}{\tikz{\draw[->](0,0)--(#1,0);}}
}
\usepackage{arydshln}

\usepackage[breaklinks=true,bookmarks=false]{hyperref}

\iccvfinalcopy 

\def\iccvPaperID{****} 

\ificcvfinal\fi

\begin{document}

\title{A CLIP-Enhanced Method for Video-Language Understanding}

\author{Guohao Li \quad Feng He \quad Zhifan Feng \\
Baidu, Beijing \\
{\tt\small \{liguohao,hefeng07,fengzhifan\}@baidu.com}
}

\maketitle
\ificcvfinal\thispagestyle{empty}\fi

\begin{abstract}
  
  This technical report summarizes our method for the Video-And-Language Understanding Evaluation (VALUE) challenge~\footnote{https://value-benchmark.github.io/challenge\_2021.html}.
  We propose a CLIP-Enhanced method to incorporate the image-text pretrained knowledge into downstream video-text tasks.
  Combined with several other improved designs, our method outperforms the state-of-the-art by $2.4\%$ ($57.58$\Arrow{.2cm}$60.00$) Meta-Ave score on VALUE benchmark.

\end{abstract}

\section{Introduction}
\label{label:intro}

Video-Language understanding attracts increasing attention in the research community~\cite{sun2019videobert,miech2019howto100m,luo2020univl,li2020hero,zhu2020actbert,lei2021less,bain2021frozen}.
Recently, Li~\cite{li2021value} proposed the Video-And-Language Understanding Evaluation (VALUE) benchmark, a unified benchmark consisting of 3 types of tasks (VideoQA, Retrieval, Captioning) and 11 datasets.
The diverse video domains and task types make it a very challenging benchmark.

Inspired by the rapid progress of large-scale image-text pre-training (e.g., CLIP~\cite{radford2021learning}), we believe the knowledge learned from image-text pairs would be helpful for video-text tasks.
Several pioneering works~\cite{portillo2021straightforward,luo2021clip4clip,fang2021clip2video} utilize the pre-trained CLIP model ~\cite{radford2021learning} and demonstrate state-of-the-art performances for several text-video retrieval tasks.
However, these existing works are specially designed for the retrieval task, thus cannot adapt to other kinds of tasks.

We incorporate the pre-trained image-text knowledge (i.e., CLIP~\cite{radford2021learning}) into a task-agnostic framework (i.e., HERO~\cite{li2020hero}), and achieve significant improvements on various downstream tasks (e.g., retrieval, captioning).
Combined with several tricks, we propose a mixed strategy for the VALUE benchmark, outperforming the baselines by $2.4\%$ ($57.58$\Arrow{.2cm}$60.00$) Meta-Ave score.

\begin{figure}
  \centering
  \begin{tabular}{cccc}
    \includegraphics[width=.45\textwidth]{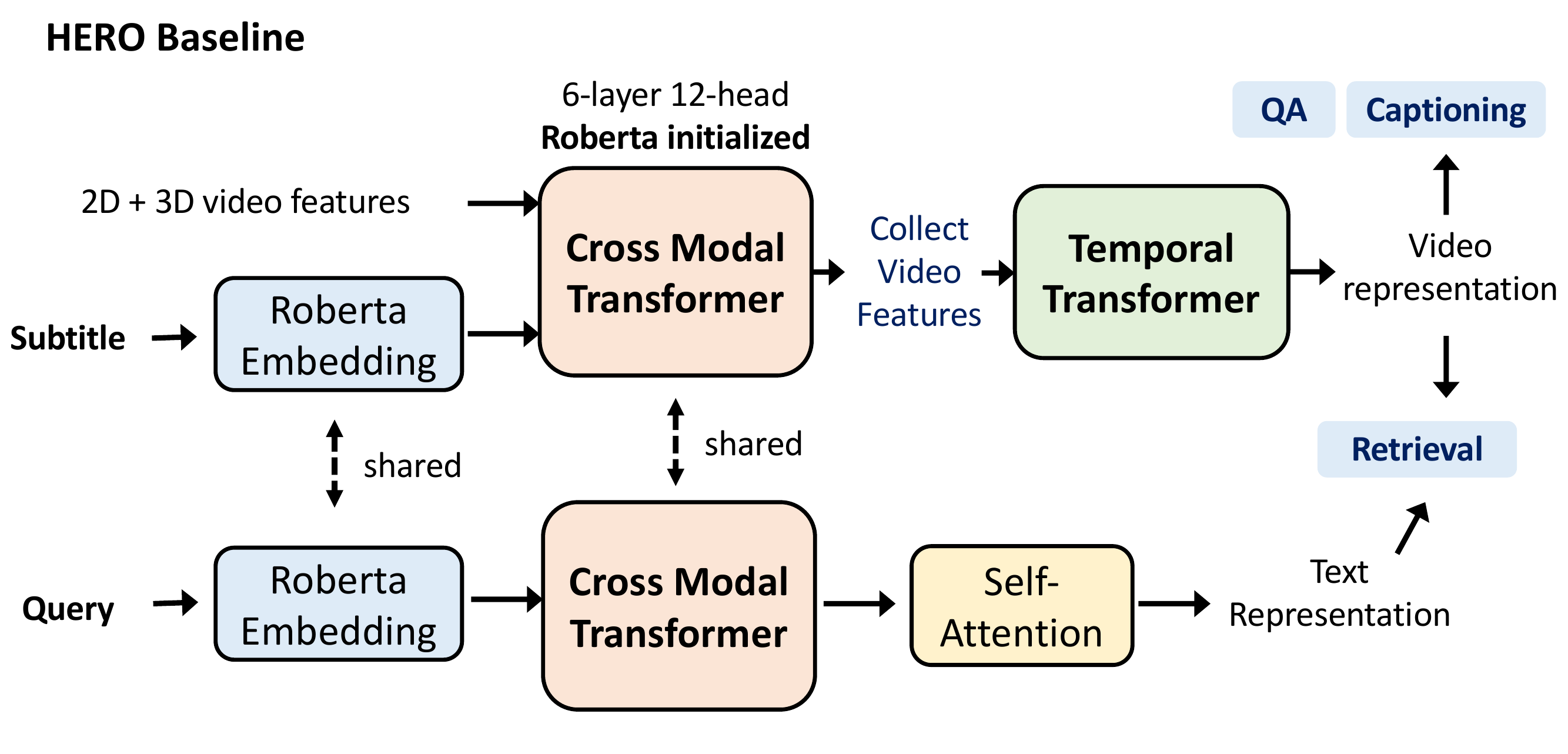} \\
    \vspace{0.5cm}
    (a) HERO baseline model. \\
    \includegraphics[width=.45\textwidth]{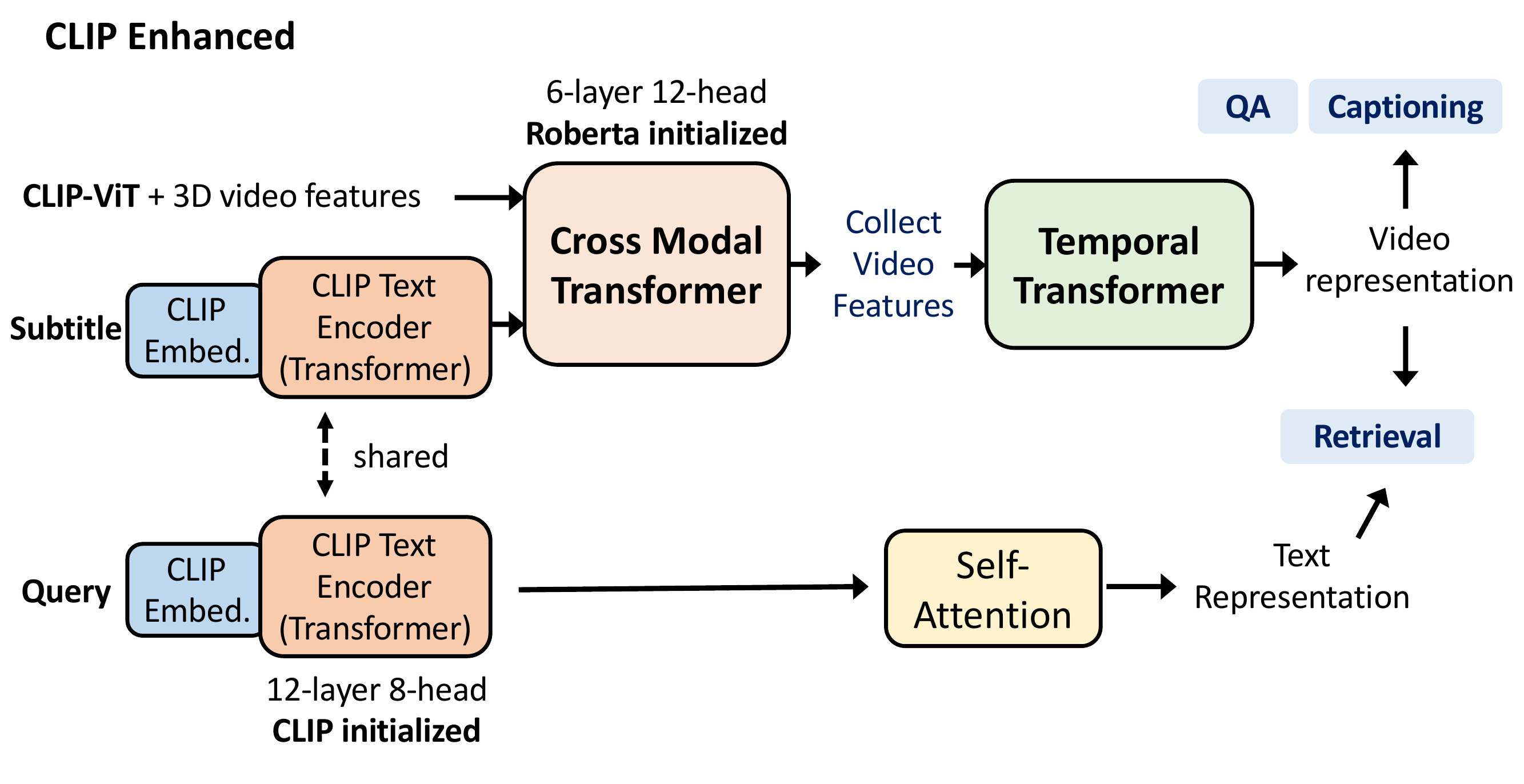} \\
    (b) Our CLIP-Enhanced model.
  \end{tabular}
  \caption{HERO baseline v.s. Our CLIP-Enhanced approach.}
  \label{fig:model_comparison}
\end{figure}

Briefly, our strategy differs from the HERO baseline~\cite{li2020hero} in two aspects:
1) The model architecture is modified to incorporate CLIP knowledge, as shown in Fig.~\ref{fig:model_comparison};
2) We use slightly different finetuning settings for different downstream tasks.


\begin{table*}[ht]
  \caption{
    Results of our method (single-model) and state-of-the-art baselines with CLIP-ViT+SlowFast features on {\bf Test (leaderboard) set}. The baselines are reported in ~\cite{li2021value}, including the following model training settings: single-task training (ST), all-task training then ST (AT $\rightarrow$ ST). The best performances are highlighted in bold.
  }
  \vspace{2mm}
  \tablestyle{2pt}{1.1} 
  \def\w{25pt} 
  \resizebox{\textwidth}{!}{
    \begin{tabular}{R{0.02\textwidth}l|R{\w}R{\w}R{\w}r|R{\w}R{\w}R{\w}R{\w}|R{\w}R{\w}r|R{\w}}
      & \multirow{2}{*}{ \newcelll{Methods}} & TVR & How2R & YC2R & \newcellr{VATEX-\\EN-R}&  TVQA   & \newcellr{How2-\\QA} & \newcellr{VIO-\\LIN} & VLEP  & TVC & YC2C & \newcellr{VATEX-\\EN-C}& \multirow{2}{*}{\newcellr{Meta-\\Ave}}\\  
      \cline{3-13}
      & & AveR & AveR & AveR & AveR & Acc. & Acc.& Acc. & Acc. & C & C & C & \\
      \shline
      & Human & - & - & - & - & 89.41 & 90.32 & 91.39  & 90.50 &  62.89 & - & 62.66 & - \\
      \shline
      & \multicolumn{5}{l}{\textit{Finetune-only}}\\
      \hline
      & ST & 8.81 & 2.13 & 42.37 & 47.02 & 71.35  & 69.59 & 64.30 & 56.77 & 50.30 & 109.89 & 55.98 & 52.59\\
      & AT\Arrow{.2cm}ST & 12.40 & 3.61 & 50.93 & \bf 49.91 & 74.38  & 71.88 & 66.80 & 68.68 & 49.41 & 110.63 & 58.09 & 56.07\\
      \shline
      & \multicolumn{5}{l}{\textit{Pretrain on ResNet+Slowfast, then Finetune}}\\
      \hline
      & ST & \bf 13.70 & 3.38 & 56.59 & 46.66 & 74.52  & 73.82 & 64.19 & 67.10 & 51.04 & 120.22 & 55.30 & 56.96 \\
      & AT\Arrow{.2cm}ST& 13.56 & 3.95 & 54.28 & 49.09 & 74.83 & \bf 74.60 & 67.18 & \bf 69.37 & 48.13 & 121.89 & 56.54 & 57.58\\
      \shline
      & \multicolumn{5}{l}{\textit{Our mixed strategy}}\\
      \hline
      & Ours & 13.12 & \bf 4.64 & \bf 62.68 & 49.86 & \bf 75.45 & 73.92 & \bf 67.47 & 68.37 & \bf 53.34 & \bf 128.87 & \bf 62.30 & \bf 60.00 \\
    \end{tabular}
  }
  \vspace{-2mm}
  \label{tab:leaderboard}
\end{table*}

\section{Method}
\label{label:method}

Our method is built based on HERO~\cite{li2020hero}, the state-of-the-art model on VALUE benchmark.
We first briefly describe the HERO method, then introduce our improved designs. 

\subsection{Baseline Method}

As shown in Fig.~\ref{fig:model_comparison} (a), HERO consists of three core components: 1) an embedding layer for textual input; 2) a Cross-Modal Transformer for video-subtitle multi-modal fusion and query representation; 3) a Temporal Transformer for learning contextualized video representation from collected video features.

The HERO framework support VideoQA, Retrieval and Captioning tasks.
Please refer to~\cite{li2020hero} for more information.

\subsection{Improved Designs}
\label{label:improved_designs}

In principle, we follow the HERO architecture for all tasks except VATEX-EN-R and VATEX-EN-C.
For the VATEX tasks, we build our CLIP-Enhanced model by replacing the default Roberta~\cite{liu2019roberta} text embedding layer with a CLIP~\cite{radford2021learning} text encoder (a 12-layer 8-head transformer, initialized with the CLIP pre-trained weights), as shown in Fig.~\ref{fig:model_comparison} (b).

We use slightly different settings when finetuning different tasks:
{\bf 1)} For the QA tasks, we adopt the all-task training (AT) setting; for other tasks, we adopt the single-task training (ST) setting.
{\bf 2)} We use resnet+slowfast features for yc2r, yc2c, how2r tasks, while for other tasks we use clip-vit+slowfast features.
These visual features are officially provided by the VALUE challenge.
{\bf 3)} For yc2r, yc2c, tvc tasks, we use training-set and validation-set data for finetuning.
{\bf 4)} We initialize the model with the HERO pre-trained weights for all tasks except the CLIP-Enhanced setting (i.e., the VATEX-EN-R and VATEX-EN-C tasks).

We didn't use extra data or features during pre-training or finetuning.
We didn't use ensemble techniques.

\section{Experiments}
\label{label:expr}

We compare our method with the state-of-the-art baselines on VALUE benchmark in Table~\ref{tab:leaderboard} and Table~\ref{tab:vatex}.

\subsection{Results on Test (leaderboard) Set}

Applying all of the improved designs as described in Sec.~\ref{label:improved_designs}, our mixed strategy achieves significant improvement compared with the baselines on Test (leaderboard) set ($60.00$ v.s. $57.58$), as shown in Table~\ref{tab:leaderboard}.

\subsection{Analysis of our CLIP-Enhanced Strategy}

In order to evaluate the effects of our CLIP-Enhanced strategy, we compare our method with the state-of-the-art baselines on VATEX-EN-R and VATEX-EN-C validation-set, as shown in Table~\ref{tab:vatex}.

\begin{table}[h]
  \centering
  \caption{Results for VATEX-EN-R and VATEX-EN-C tasks with CLIP-ViT+SlowFast features on the {\bf Validation set}. The baseline results are reported in ~\cite{li2021value}. Note that the scores marked with star (*) are unfairly high because of data leakage.}
  \begin{tabular}{c|cc}
    \quad Methods \quad & VATEX-EN-R & VATEX-EN-C \\
    \shline
    \multicolumn{3}{l}{\textit{Finetune-only}}\\
    \hline
    ST & 65.62 & 56.97 \\ 
    AT\Arrow{.2cm}ST & 78.72* & 58.13\\ 
    \shline
    \multicolumn{3}{l}{\textit{Pretrain, then Finetune}} \\
    \hline
    ST & 66.49 & 56.19 \\ 
    AT\Arrow{.2cm}ST & 79.95*  & 56.35 \\ 
    \shline
    \multicolumn{3}{l}{\textit{Our strategy: CLIP-Enhanced, Finetune-only}} \\
    \hline
    Ours (ST) & {\bf 68.68} & {\bf 61.36} \\
  \end{tabular}
  \label{tab:vatex}
\end{table}

Our CLIP-Enhanced method achieves best performance except for the AT\Arrow{.2cm}ST baselines on the VATEX-EN-R task.
However, after checking the training details, we find that the AT\Arrow{.2cm}ST improvments are indeed from data leakage during All-Task training (AT), i.e., the validation samples for VATEX-EN-R task are accidentally included in the training set of VATEX-EN-C tasks.

Leaving out the unfairly high scores (marked with *), our CLIP-Enhanced method achieves best performance with obvious improvement over the baselines ($2\%$ for VATEX-EN-R, $3\%$ for VATEX-EN-C).

Nevertheless, we observe that our CLIP-Enhanced method fails for other type of datasets (e.g., how2, tv).
The main reason seems to be that -- the how2 or tv datasets are quite different from the image-text pairs that the CLIP model is pre-trained on.

{\small
  \bibliographystyle{ieee_fullname}
  \bibliography{references}
}

\end{document}